\documentclass[usenames,dvipsnames]{article} 
\usepackage{iclr2024_conference,times}

\usepackage{amsmath}
\usepackage{amsthm}
\usepackage{amsfonts}

\usepackage{bm}
\usepackage{xcolor}

\usepackage{url}
\usepackage{algorithm}
\usepackage[noend]{algorithmic}
\usepackage{graphicx}
\renewcommand{\algorithmiccomment}[1]{\bgroup\hfill $\triangleright$ ~#1\egroup}
\usepackage{hyperref}
\usepackage{amsmath}
\usepackage{caption}

\title{Weight Norm Control}

\def\R{{\rm I\hspace{-0.50ex}R}}
\def\E{\mathds{E}}
\def\D{{\mathcal{D}}}
\def\H{\textbf{H}}
\def\A{\cal{A}}
\newcommand{\vc}[1]{\boldsymbol{#1}}

\iclrfinalcopy

\author{Ilya Loshchilov\\
\texttt{ilya.loshchilov@gmail.com} 
}

\begin{document}

\maketitle

\begin{abstract} 

We note that decoupled weight decay regularization is a particular case of weight norm control where the target norm of weights is set to 0. Any optimization method (e.g., Adam) which uses decoupled weight decay regularization (respectively, AdamW) can be viewed as a particular case of a more general algorithm with weight norm control (respectively, AdamWN). We argue that setting the target norm of weights to 0 can be suboptimal and other target norm values can be considered. For instance, any training run where AdamW achieves a particular norm of weights can be challenged by AdamWN scheduled to achieve a comparable norm of weights. We discuss various implications of introducing weight norm control instead of weight decay.   

\end{abstract}

\section{Introduction}

\def\R{{\rm I\hspace{-0.50ex}R}}
\def\E{\mathds{E}}
\def\D{{\mathcal{D}}}
\def\H{\textbf{H}}
\def\A{\cal{A}}

\newcommand{\ma}[1]{\mathchoice{\mbox{\boldmath$\displaystyle#1$}}
  {\mbox{\boldmath$\textstyle#1$}} {\mbox{\boldmath$\scriptstyle#1$}}
  {\mbox{\boldmath$\scriptscriptstyle#1$}}}
\renewcommand{\ma}[1]{\mathnormal{\mathbf{#1}}}
\newcommand{\mstr}[1]{\mathrm{#1}}
\newcommand{\C}{ \ensuremath{\ma{C}} }
\newcommand{\I}{ \ensuremath{\ma{I}} }
\newcommand{\M}{ \ensuremath{\ma{M}} }
\newcommand{\NormalNullC}{{\mathcal N}  \hspace{-0.13em}\left({\ma{0},\C\,}\right)}
\newcommand{\dd}{n}

\def\UU{{\rm I\hspace{-0.50ex}U}}

\def\RR{{\rm I\hspace{-0.50ex}R}}

\def\NormOI{{\mathcal N}  \hspace{-0.13em}\left({\ma{0}, \ensuremath{\ma{I}}\,}\right)}
\def\ONE{{\rm 1\hspace{-0.80ex}1}}
\def\Id{\ensuremath{\ma{I}}}
\def\MYUNDERLINE{ $\noindent\underline{\makebox[0.06in][l]{}}$ }
\def\x{\bm{\theta}}
\def\y{\vc{y}}
\def\m{\vc{m}}
\def\vy{\vc{y}}
\newcommand{\HYP}{H}
\def\UU{{\rm I\hspace{-0.60ex}U}}

\definecolor{newcolor}{rgb}{0.8,1,1}

\newcommand{\adamcolor}{Thistle}
\newcommand{\ouradamcolor}{SpringGreen}
\newcommand{\adam}[1]{\colorbox{\adamcolor}{$\displaystyle #1$}}
\newcommand{\adamtext}[1]{\colorbox{\adamcolor}{#1}}
\newcommand{\ouradam}[1]{\colorbox{\ouradamcolor}{$\displaystyle #1$}}
\newcommand{\ouradamtext}[1]{\colorbox{\ouradamcolor}{#1}}

In the weight decay described by \citet{hanson1988comparing}, the weights $\bm{\theta}$  decay exponentially as 
\begin{eqnarray}
	\bm{\theta}_{t+1} = (1 - \lambda) \bm{\theta}_t - \alpha \nabla f_t(\bm{\theta}_t), \label{eq:wdecay}
\end{eqnarray}
where $\lambda$ defines the rate of the weight decay per step and $\nabla f_t(\bm{\theta}_t)$ is the $t$-th batch gradient to be multiplied by a learning rate $\alpha$. More recently, \cite{loshchilov2019decoupled} demonstrated that this original formulation is not equivalent to standard L$_2$ regularization when used in adaptive gradients methods such as Adam~\citep{kingma2014adam}. They introduced AdamW algorithm where Adam's loss-based update is decoupled from weight decay. 

\section{Weight norm control}

Since the loss-based and weights-based updates of Equation \ref{eq:wdecay} are decoupled, the weight decay is

\begin{eqnarray}
	\bm{\theta}_{t+1} = \bm{\theta}_t - \lambda \bm{\theta}_t \label{eq:wdecayonly}
\end{eqnarray}

One can view it as a particular case of 

\begin{eqnarray}
	\bm{\theta}_{t+1} = \bm{\theta}_{t}  - k_t \left( 1 - \frac{r_t \| \bm{\theta}_0 \|}{\| \bm{\theta}_t \|} \right) \bm{\theta}_{t}, \label{eq:wcontrol}
\end{eqnarray}

when the ratio between target and initial norms of weights $r_t=0$, and the update rate $k_t=\lambda$. $k_t$ is called update rate and not weight decay because for $r_t>1$, the weights should increase with rate $k_t$ and not decrease/decay. When $k_t=1$, the norm of weights $\| \bm{\theta}_t \|$ is immediately updated to its target value $r_t \| \bm{\theta}_0 \|$. It is important to note that in case if we consider different initialization scenarios (and thus modify $\| \bm{\theta}_0 \|$), we can define the target weight norm not w.r.t. $\bm{\theta}_0$ but in absolute/raw values. The use of $r_t$ as the ratio is primarily for practical reasons. 

\begin{figure*}[t]
\begin{center}
    \includegraphics[width=0.5\textwidth]{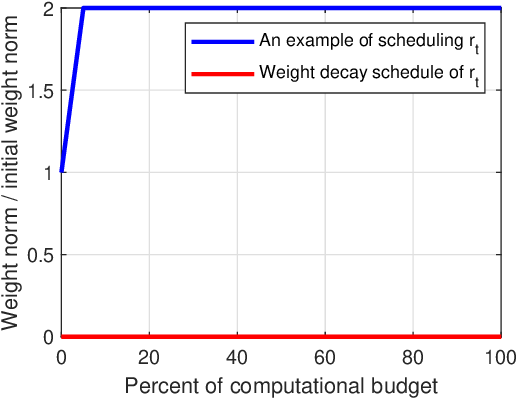} 
\caption{\label{fig:rtschedule} An example of scheduling $r_t$ (blue line). Weight decay corresponds to $r_t=0$ (red line). }
\vspace*{-0.25cm}
\end{center}
\end{figure*}

Figure \ref{fig:rtschedule} shows $r_t$ over time when weight decay is considered (red line). It also shows an example of scheduling $r_t$ where the weight norm slowly grows  to double its initial value and then remain constant.

\newcounter{SubState}
\renewcommand{\theSubState}{\arabic{ALC@line}.\alph{SubState}}

\newcommand{\mysubstate}[1]{%
    \refstepcounter{SubState}
    \item[\theSubState] #1
}

\begin{algorithm}[tb!]
\caption{AdamW and \ouradamtext{AdamWN}}
\footnotesize
\label{algo_adam}
\begin{algorithmic}[1]
\STATE{\textbf{given} $\alpha = 0.001, \beta_1 = 0.9, \beta_2 =0.999, \epsilon = 10^{-8}, \lambda\in \R$} \label{adam-Given}
\STATE{\textbf{initialize} time step $t \leftarrow 0$, parameter vector $\bm{\theta}_{t=0} \in \R^n$,  first moment vector $\vc{m}_{t=0} \leftarrow \vc{0}$, second moment vector  $\vc{v}_{t=0} \leftarrow \vc{0}$,  target weight norm ratio \ouradam{r_t \in \R} where $r_t \| \bm{\theta}_0 \|$ is the target weight norm for ${\theta}_t$, weight norm update rate \ouradam{k_t \in [0,1]}  }
\REPEAT
	\STATE{$t \leftarrow t + 1$}
	\STATE{$\nabla f_t(\bm{\theta}_{t-1}) \leftarrow  \text{SelectBatch}(\bm{\theta}_{t-1})$}  \COMMENT{select batch and return the corresponding gradient}
	\STATE{$\vc{g}_t \leftarrow \nabla f_t(\bm{\theta}_{t-1})$  }
	\STATE{$\vc{m}_t \leftarrow \beta_1 \vc{m}_{t-1} + (1 - \beta_1) \vc{g}_t $} \label{adam-mom1} \COMMENT{here and below all operations are element-wise}
	\STATE{$\vc{v}_t \leftarrow \beta_2 \vc{v}_{t-1} + (1 - \beta_2) \vc{g}^2_t $} \label{adam-mom2}
	\STATE{$\hat{\vc{m}}_t \leftarrow \vc{m}_t/(1 - \beta_1^t) $} \COMMENT{$\beta_1$ is taken to the power of $t$} \label{adam-corr1}
	\STATE{$\hat{\vc{{v}}}_t \leftarrow \vc{v}_t/(1 - \beta_2^t) $} \COMMENT{$\beta_2$ is taken to the power of $t$} \label{adam-corr2}
	\STATE{$\eta_t \leftarrow \text{SetScheduleMultiplier}(t)$}	\COMMENT{can be fixed, decay, or also be used for warm restarts}
       \STATE{$\bm{\theta}_t \leftarrow \bm{\theta}_{t-1} - \eta_t \ \alpha  \hat{\vc{m}}_t / (\sqrt{\hat{\vc{v}}_t} + \epsilon)$} \label{adam-xupdate}
    \STATE{ \textbf{Regularization}: }
\setcounter{SubState}{0}

\mysubstate{$\bm{\theta}_t \leftarrow \bm{\theta}_{t} - \eta_t \alpha_0 \lambda\bm{\theta}_{t}$} \COMMENT{common implementation of decoupled weight decay}

\mysubstate{$\bm{\theta}_t \leftarrow \bm{\theta}_{t} - \eta_t \lambda\bm{\theta}_{t} $} 
\COMMENT{implementation of decoupled weight decay as in AdamW paper}

\mysubstate{\ouradam{\bm{\theta}_t \leftarrow \bm{\theta}_{t}  - k_t \left(1 - \frac{r_t \| \bm{\theta}_0 \|}{\| \bm{\theta}_t \|} \right) \bm{\theta}_{t} }}
\COMMENT{proposed $r_t$-based scheduling of weight norm}

\item[] Notes: 13.a is a particular case of 13.c when $r_t=0$ and $k_t=\eta_t \alpha_0 \lambda$ \\ 
\hspace{2.615em}  13.b  is a particular case of 13.c when $r_t=0$ and $k_t=\eta_t \lambda$
\UNTIL{ \textit{stopping criterion is met} }
\RETURN{optimized parameters $\bm{\theta}_t$}
\end{algorithmic}
\end{algorithm}

\begin{figure*}[t]
\begin{center}
    \includegraphics[width=1\textwidth]{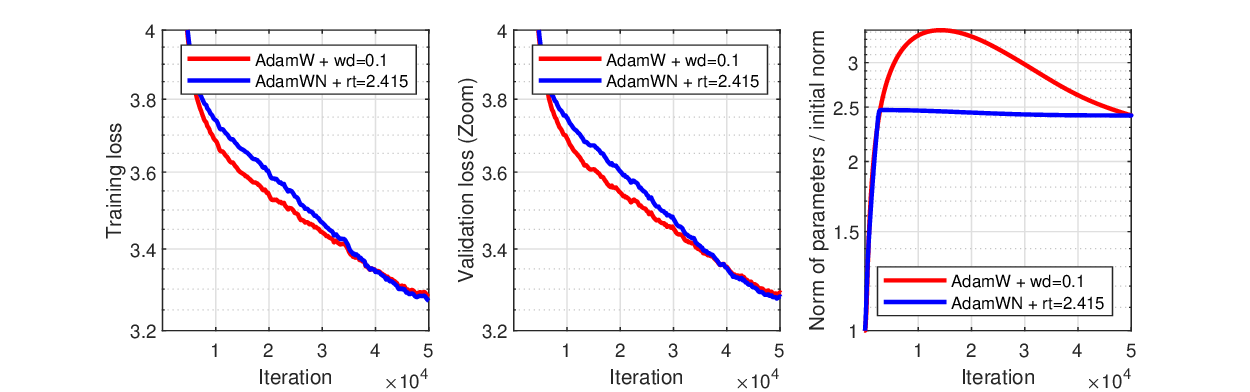} 
\caption{\label{fig:50k} AdamW (red) with weight decay 0.1 and AdamWN (blue) with $r_t=2.415$ which corresponds to the final weight norm obtained by AdamW. These experiments correspond to batchsize of only 65k tokens. }
\vspace*{-0.25cm}
\end{center}
\end{figure*}

\begin{figure*}[t]
\begin{center}
    \includegraphics[width=1\textwidth]{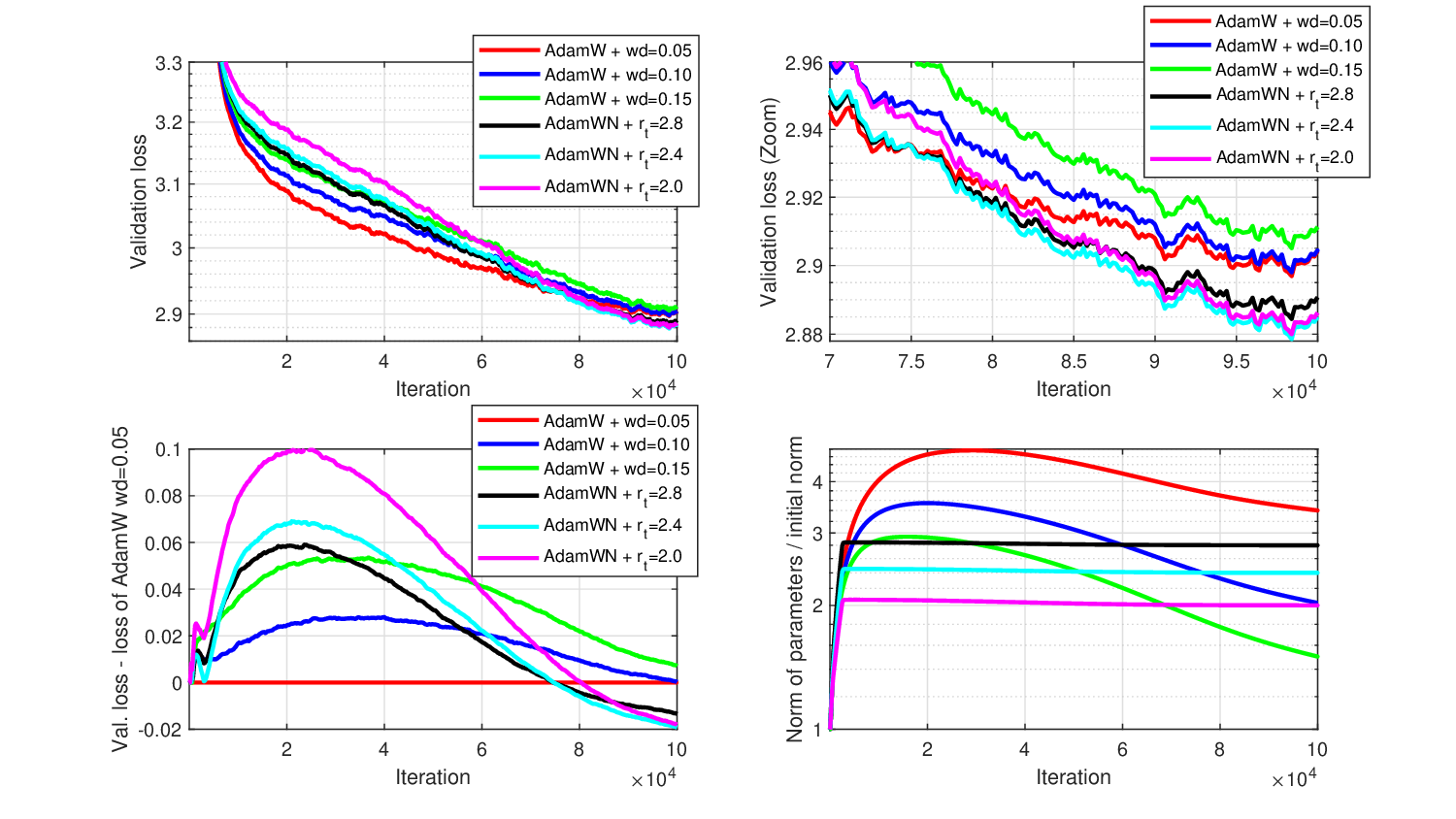} 
\caption{\label{fig:100k} AdamW with different settings of weight decay and AdamWN with different settings of target weight norm. These experiments correspond to batchsize of 500k tokens. Bottom-Left subfigure shows validation loss w.r.t. validation loss of AdamW with weight decay factor set to 0.05. }
\vspace*{-0.25cm}
\end{center}
\end{figure*}

\section{AdamWN: Adam with Weight Norm Control}

We propose AdamWN as a version of Adam algorithm where the weight norm is controlled according to Equation \eqref{eq:wcontrol}. 
Algorithm \ref{algo_adam} shows the original AdamW. The modifications corresponding to AdamWN are depicted by green color. The difference between algorithms appears in line 13 where regularization is performed. Most implementations of AdamW correspond to line 13.a where weights are decayed by a factor $\eta_t \alpha_0 \lambda$ and thus weight decay factor $\lambda$ is not decoupled from learning rates $\alpha_0$ scheduled by $\eta_t$ (e.g., cosine annealing schedule). In practice, this complicates hyperparameter tuning because, as the original paper of AdamW demonstrated, the numerical value of weight decay is typically in order of $\alpha_0 \lambda$. Line 13.b corresponds to weight decay as proposed in the original AdamW paper where the initial learning rate $\alpha_0$ and weight decay $\lambda$ are decoupled. 

Our proposed weight norm control is shown in line 13.c. To easier understand this equation one should consider particular settings of $k_t$ and $r_t$. If $k_t=1$, then $\bm{\theta}_t$ will be immediately updated to $r_t \bm{\theta}_0$, i.e., all weights will be rescaled by $r_t$. If $k_t$ is some small value, then the weights will be slowly (with rate $k_t$) updated towards $r_t \bm{\theta}_0$. In a different particular situation, when $r_t=0$ and $k_t=\eta_t \alpha_0 \lambda$, we recover the common update of AdamW given in line 13.a. Also, when $r_t=0$ and $k_t=\eta_t \lambda$, we recover the original update of AdamW given in line 13.b. 

Weight norm control can be used with any optimization algorithm, here we used AdamW because of its popularity and to demonstrate that AdamW is a particular case of AdamWN.

\section{Experimental Results}

We use NanoGPT framework \citep{karpathy2023nanogpt} to train GPT-2 \citep{radford2019language} on OpenWebText \citep{Gokaslan2019OpenWeb}. The framework reproduces GPT-2 experiments for different model sizes. In all experiments, we consider GPT-2-small model with 124M parameters resulting from $n_{layer}$=12, $n_{head}$=12, $n_{embd}$=768. If not mentioned otherwise, all other hyperparameters are set as in the original NanoGPT implementation. We use RTX 4090 GPUs where 100k iterations of training with 491k tokens per batch take 5 days (such experiments are shown in Figure \ref{fig:100k}). In all experiments, we set the initial learning rate $\alpha_0$ to 1e-3 and use cosine annealing from 1.0 to 0.1 as suggested for NanoGPT experiments. When measuring norm of weights we do not include parameters of LayerNorm \citep{ba2016layer} since they are also not weight decayed in the original NanoGPT implementation. However, our initial experiments with up to 200k iterations (10 days of compute) involved weight control of LayerNorm parameters (see Appendix \ref{sec:layernormcontrol}). 

Figure \ref{fig:50k} shows our initial experiments with AdamW and AdamWN with small batch size of only 65k tokens. First, we launch AdamW with weight decay factor set to 0.1 and measure the $L_2$-norm of weights w.r.t. their initial norm. When measured at the last iteration, this value equals 2.415. Then, we schedule the weight norm in AdamWN to achieve that $r_t$ by linearly increasing it from 1.0 to 2.415 within the first 2500 iterations. After that, $r_t$ remains constant. It should be noted that while $r_t$ defines the target ratio of the weight norm w.r.t. its initial norm at $t=0$, the actual weight norm can deviate from that target value. The value of $k_t$ from Equation \eqref{eq:wcontrol} controls how much deviation is allowed. In all experiments, we set $k_t$ to 1e-2 as this allows the weight norm to slightly deviate from its target. We do not observe the algorithm to be very sensitive to this hyperparameter. Figure \ref{fig:50k} shows that AdamWN can achieve slightly better training and validation error values than AdamW while following a predefined schedule of $r_t$. 

Figure \ref{fig:100k} shows AdamW with different settings of weight decay $\lambda \in [0.05; 0.10; 0.15]$ and AdamWN with different settings of final values of $r_t \in [2.0; 2.4; 2.8]$. These experiments are performed for 100k iterations and about 500k tokens per batch and thus are much longer than 50k iterations experiments with 65k tokens per batch shown in Figure \ref{fig:50k}. The longer runs allow to demonstrate a more significant difference between AdamW and AdamWN. More specifically, AdamWN outperforms AdamW for comparable settings of the final weight norm. Figure \ref{fig:100k} Bottom-Left better illustrates that AdamWN runs are initially slower due to the weight norm constraint. However, they perform better at later stages of optimization. Notably, AdamW with 0.1 weight decay and AdamWN with final $r_t=2.0$ achieve the same weight norm but AdamWN achieves validation loss of 2.884 while AdamW achieves 2.902. This seemingly small difference is actually quite substantial: Figure \ref{fig:200k} of Appendix A shows that AdamW with 0.1 weight decay achieves validation loss of 2.88 after performing 200k (2 times more) iterations. We suspect that the gap in performance between AdamWN and AdamW will grow with computational budget.   

A possible explanation of our observations could be that AdamW with weight decay searches on different scales and wastes some computational resources while scaling-up based on loss-based gradients and then scaling-down based on weight decay. Instead, AdamWN spends most computational resources by searching closely within the target and final weight norm. 

\section{Credit assignment}

Weight norm control is a very simple idea. It can be assumed that in the context of neural networks it was discussed before \citet{hanson1988comparing} while  \citet{hanson1988comparing} adopted weight decay as a particular case of weight norm control. It can be safely assumed that in more general context of L$_2$ regularization, machine learning and optimization the idea was first discussed many decades ago. 

More recently, in deep learning era, the idea to fix weight  norm ($r_t=1$) or adjust algorithm parameters based on weight norm is studied in various works (see, e.g., weight normalization via a reparameterization of weight vectors by \cite{NIPS2016_ed265bc9}). To the best of our knowledge, the idea of scheduling $r_t$ and govern it by Equation \eqref{eq:wcontrol} is novel and 
is not used by practitioners.

The first version of this paper was submitted to arXiv on 19 November 2023, 12 hours after we received  a Google Scholar notification about a relevant arXiv paper by \cite{Franke2023} submitted on 15 November 2023 and announced by arXiv on 16 or 17 of November 2023. The emergence of this work prompted us to expedite the completion and submission of our paper, affirming that our research was conducted in parallel and independently of \cite{Franke2023} (also, we did not have any research-related communication with the authors). Experiments with 100k iterations shown in Figure \ref{fig:100k}  take 5 days to perform, while experiments with 200k iterations shown in Figure \ref{fig:200k} of Appendix A take 10 days to perform on RTX 4090 GPUs used in this work. 

\cite{Franke2023} proposes Constrained Parameter Regularization (CPR) where ``Instead of applying a constant penalty uniformly to all parameters, we enforce an upper bound on a statistical measure (e.g., the $L_2$-norm) of individual parameter groups". Already from this description one can see that our work differs because CPR enforces an upper bound on norm of weights (or other statistical measure)  while our approach constrains the weight norm to have a specific target value. In other words, CPR constrains solutions to be anywhere inside a hyper-ellipsoid with a particular $L_2$-norm while our approach suggests to constrain solutions to lie on the  surface of a hyper-ellisoid with a particular target $L_2$-norm. 
In this sense, CPR is closely related to max-norm regularization \citep{srivastava2014dropout} where "a constraint was imposed during optimization by projecting weights $w$ onto the surface of a ball of radius $c$, whenever $w$ went out of it''. While both max-norm and CPR can be modified to correspond to our approach, without such modifications they are just two related approaches. The reason we mention CPR here and submitted our preprint earlier than planned is because our approach described by Equation \eqref{eq:wcontrol} is so trivial that only a few modifications are needed for CPR (as well as for weight decay) to be converted to our method. For instance, replacing less-or-equal constraint to equal constraint would make CPR more similar to our approach. An additional necessary modification is scheduling of $r_t$. In CPR (its Kappa-Is variant), the maximum weight norm that is used as a constraint is set to be the norm of weights measured right  after a given number $\kappa$ of iterations. Therefore, in contrast to $r_t$ schedule which is set before running the algorithm, the maximum weight norm used in the Kappa-Is variant of CPR is not decoupled from the optimization algorithm and is a function of algorithm's internals, e.g., learning rate values which affect the weight norm after $\kappa$ iterations (this ``seemingly linear dependence'' w.r.t. the learning rate was also noted by the authors of CPR). 

\section{Discussion}

Weight norm control represents a more general case of weight decay where the $L_2$-norm of weights does not have to decay to zero as in weight decay but can be adapted or scheduled. 
Scheduling the weight norm to reach a particular value $r_t \| \bm{\theta}_0 \|$ forces the algorithm to search on the surface of hyper-ellipsoids with the corresponding  target $L_2$-norm. The algorithm can deviate from that target norm due to loss-based gradients. The amplitude of  such deviations is affected by $k_t$. 

Importantly, the optimal weight decay factor in AdamW is a function of the total number of iterations (among other hyperparameters) because the longer we run AdamW the more it will decay the weight norm (as much as loss-based gradient-based steps it). This and the fact that the final weight norm is sensitive to raw values of weight decay (see Figure \ref{fig:100k}-Bottom-Right) makes it difficult to select the optimal weight decay factor.  Also, given that most implementations of AdamW do not decouple learning rate and weight decay, any change of the initial learning rate or its online adaptation will affect the effective rate of weight decay and thus the final weight norm (in contrast to our approach where it is predefined by $r_t$). 



\bibliography{iclr2024_conference}
\bibliographystyle{iclr2024_conference}

\cleardoublepage
\setcounter{page}{1}
\setcounter{section}{0}

\appendix
{\begin{center}\Large{\textbf{Appendix}}\end{center}}
\section{Initial experiments where norms of LayerNorm parameters are controlled}
\label{sec:layernormcontrol}

\begin{figure*}[t]
\begin{center}
    \includegraphics[width=1\textwidth]{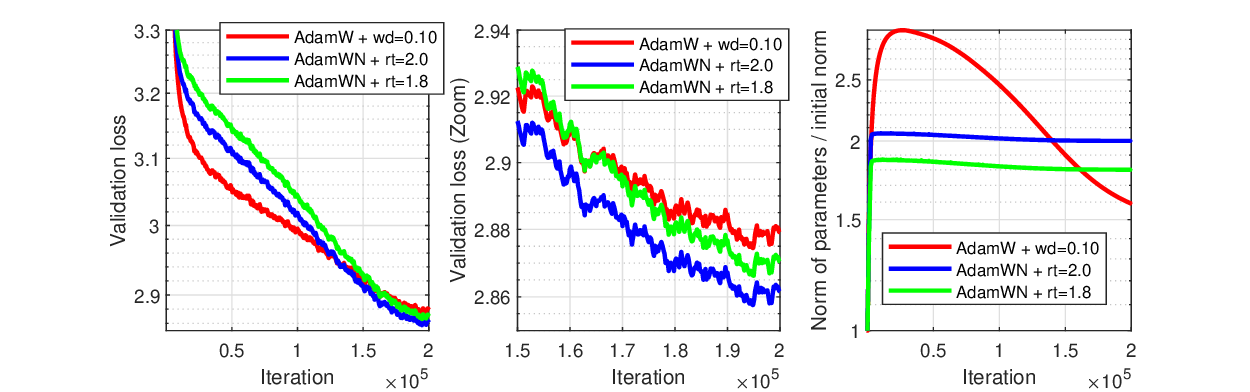} 
\caption{\label{fig:200k} AdamW (red) with weight decay 0.1 and AdamWN with final $r_t=2.0$ (blue) and $r_t=1.8$ (green). }
\vspace*{-0.25cm}
\end{center}
\end{figure*}

In our main paper we discuss experiments where LayerNorm parameters are not weight decayed or weight controlled because they are not weight decayed in the GPT-2 implementation of NanoGPT. However, our initial experiments involved weight control of LayerNorm parameters. Figure \ref{fig:200k} shows that AdamWN (where all parameters are weight norm controlled) with $r_t=2.0$ and $r_t=1.8$ outperforms AdamW (where all except LayerNorm parameters are weight decayed) with weight decay 0.1 after 200k iterations (10 days of compute). After performing these experiments, we noted that better performance can be achieved without weight controlling LayerNorms parameters as NanoGPT suggests to do when weight decay is applied. Therefore, all experiments in the main text of the paper do not consider weight decay/control of LayerNorm parameters. 

\end{document}